\documentclass[11pt, a4paper]{article} 

\usepackage{inputenc} 
\usepackage[T1]{fontenc}    
\usepackage{lmodern}        
\usepackage[english]{babel} 
\usepackage{parskip}

\usepackage{geometry}       
\usepackage{microtype}      

\usepackage{amsmath}
\usepackage{amsfonts} 
\usepackage{amssymb}  
\usepackage{multicol} 
\usepackage{lipsum}   
\usepackage{caption}
\usepackage{amsmath}        
\usepackage{amssymb}        
\usepackage{amsfonts}       
\usepackage{mathtools}      
\usepackage{abstract}
\usepackage[noend]{algpseudocode}
\usepackage{algorithm}

\usepackage{graphicx}       
\graphicspath{{images/}}    
\usepackage{xcolor}         
\definecolor{myblue}{rgb}{0.2, 0.5, 0.8} 
\definecolor{mygray}{rgb}{0.95, 0.95, 0.95} 

\usepackage{algorithm}
\usepackage[noend]{algpseudocode}

\usepackage{booktabs}       
\usepackage{longtable}      
\usepackage{etoolbox}
\AtBeginEnvironment{longtable}{\footnotesize}
\usepackage{array}          

\usepackage{enumitem}       
\usepackage{tabularx}
\usepackage{booktabs} 
\usepackage{tabularx} 
\usepackage{ragged2e} 
\usepackage{listings}       
\lstdefinestyle{mystyle}{
	backgroundcolor=\color{mygray}, 
	commentstyle=\color{green!50!black},
	keywordstyle=\color{myblue},
	numberstyle=\tiny\color{gray},
	stringstyle=\color{purple!80!black},
	basicstyle=\ttfamily\footnotesize,
	breakatwhitespace=false,
	breaklines=true,
	captionpos=b,
	keepspaces=true,
	numbers=left,
	numbersep=5pt,
	showspaces=false,
	showstringspaces=false,
	showtabs=false,
	tabsize=2,
	frame=tb, 
	framesep=5pt,
	framerule=0.5pt,
	rulecolor=\color{gray!50}
}
\usepackage{fancyhdr}       
\usepackage{titlesec}
\titleformat{\section}
{\Large\bfseries\color{myblue}} 
{\thesection}                   
{1em}                           
{}                              
[\titlerule\vspace{0.5ex}]      
\titlespacing*{\section}{0pt}{3.5ex plus 1ex minus .2ex}{2.3ex plus .2ex} 

\titleformat{\subsection}
{\large\bfseries\color{myblue!80!black}} 
{\thesubsection}
{1em}
{}
\titlespacing*{\subsection}{0pt}{3.25ex plus 1ex minus .2ex}{1.5ex plus .2ex}

\titleformat{\subsubsection}
{\normalsize\bfseries} 
{\thesubsubsection}
{1em}
{}
\titlespacing*{\subsubsection}{0pt}{3.25ex plus 1ex minus .2ex}{1.5ex plus .2ex}

\usepackage{hyperref}
\hypersetup{
	colorlinks=true,            
	linkcolor=myblue,           
	citecolor=green!60!black,   
	urlcolor=blue!80!black,     
	pdftitle={\@title},         
	pdfauthor={\@author},       
	pdfsubject={A modern LaTeX article}, 
	pdfkeywords={LaTeX, article, template, modern, scientific}, 
	bookmarksopen=true,         
	bookmarksnumbered=true      
}

\usepackage[
style=nature,
backend=biber,
sorting=none,
maxnames=3,
minnames=1,
maxbibnames=99,
minbibnames=1,
giveninits=true,
terseinits=true,
uniquename=init,
uniquelist=false,
doi=true,
url=false,
isbn=false,
eprint=false
]{biblatex}

\DeclareFieldFormat{title}{\mkbibemph{#1}}
\DeclareFieldFormat[article]{title}{#1}
\DeclareFieldFormat{journaltitle}{#1}
\DeclareFieldFormat{volume}{\textbf{#1}}
\DeclareFieldFormat{pages}{#1}
\DeclareFieldFormat{doi}{\url{https://doi.org/#1}}

\DeclareNameAlias{sortname}{family-given}
\DeclareNameAlias{default}{family-given}

\renewbibmacro*{volume+number+eid}{%
	\printfield{volume}%
	\setunit*{\addcomma\space}%
	\printfield{number}%
	\setunit{\addcomma\space}%
	\printfield{eid}}

\renewbibmacro*{date}{%
	\iffieldundef{year}
	{}
	{\printtext[parens]{\printdate}}}
\usepackage{csquotes}           
\usepackage{multicol}

\makeatletter

\def\maketitle@title{%
	\thispagestyle{empty} 
	\begin{center}%
		\vspace*{1em} 
		
		\noindent 
		\parbox{0.8\textwidth}{%
			\centering 
			\rule{\linewidth}{0.6pt}
		}
		\par 
		
		\vspace{0.8em} 
		
		{\Huge \sffamily \bfseries \@title \par}
		
		\vspace{0.6em} 
		
		\noindent
		\parbox{0.8\textwidth}{%
			\centering
			\rule{\linewidth}{0.6pt}%
		}
		\par 
		
		\vskip 2.5em 
	\end{center}%
	\setlength{\AB@aftertitleskip}{0pt}%
}
\usepackage{authblk}

\makeatother

\title{
	\rule{\textwidth}{2pt} 
	\\[0.6em]
\textbf{Predicting ICU In-Hospital Mortality Using Adaptive Transformer Layer Fusion}
	\\[0.2em]
	\rule{\textwidth}{1.5pt} 
}

\author[1,2,3]{Han Wang$^{\dagger}$}
\author[1]{Ruoyun He$^{\dagger}$}
\author[1,4]{Guoguang Lao$^{\dagger}$}
\author[1]{Ting Liu$^{\dagger}$}
\author[5]{Hejiao Luo}
\author[6]{Changqi Qin}
\author[3]{Hongying Luo}
\author[3]{Junmin Huang}
\author[3]{Zihan Wei}
\author[3]{Lu Chen}
\author[3]{Yongzhi Xu}
\author[7]{Ziqian Bi}
\author[8]{Junhao Song}
\author[9]{Tianyang Wang}
\author[10]{Chia Xin Liang}
\author[11]{Xinyuan Song}
\author[3]{Huafeng Liu\thanks{hf-liu@263.net (\textit{Corresponding Author})}}
\author[3,12]{Junfeng Hao\thanks{ygzhjf85@gmail.com (\textit{Corresponding Author})}}
\author[1,3]{Chunjie Tian\thanks{tcjent@outlook.com (\textit{Corresponding Author})}}

\affil[1]{Dept. of Otorhinolaryngology, Affiliated Hospital of Guangdong Medical University, Zhanjiang 524000, China}
\affil[2]{First Clinical College, Guangdong Medical University, Zhanjiang, China}
\affil[3]{Guangdong Provincial Key Laboratory of Autophagy and Major Chronic Non-communicable Diseases; Institute of Nephrology, Affiliated Hospital of Guangdong Medical University, Zhanjiang 524001, China}
\affil[4]{The First Dongguan Affiliated Hospital, Guangdong Medical University, Dongguan 523710, China}
\affil[5]{Dept. of Critical Care Medicine, Affiliated Hospital of Guangdong Medical University, Zhanjiang, China}
\affil[6]{CICU, The Seventh Affiliated Hospital, Sun Yat-sen University, Shenzhen, China}
\affil[7]{Beijing University of Technology, Beijing 100124, China}
\affil[8]{China Agricultural University, Beijing 100083, China}
\affil[9]{Xi'an Jiaotong-Liverpool University, Suzhou 215123, China}
\affil[10]{JTB Technology Corp., Tainan 741, Taiwan}
\affil[11]{School of Physics, Peking University, Beijing 100871, China}
\affil[12]{Dept. of Family Medicine, Shengjing Hospital of China Medical University, Shenyang 110022, China}

	\date{\footnotesize  June 2025}
	
\begin{document}
		
		\maketitle 
		
		\begin{abstract}
			\linespread{1.25}\selectfont  
			Early identification of high‑risk ICU patients is crucial for directing limited medical resources. We introduce \textbf{ALFIA (Adaptive Layer Fusion with Intelligent Attention)}, a modular, attention‑based architecture that jointly trains LoRA (Low-Rank Adaptation) adapters and an adaptive layer‑weighting mechanism to fuse multi‑layer semantic features from a BERT backbone. Trained on our rigorous cw‑24 (CriticalWindow-24) benchmark, ALFIA surpasses state‑of‑the‑art tabular classifiers in AUPRC while preserving a balanced precision–recall profile. The embeddings produced by ALFIA’s fusion module, capturing both fine‑grained clinical cues and high‑level concepts, enable seamless pairing with GBDTs (CatBoost/LightGBM) as \textbf{ALFIA‑boost}, and deep neuro networks as \textbf{ALFIA-nn}, yielding additional performance gains. Our experiments confirm ALFIA’s superior early‑warning performance, by operating directly on routine clinical text, it furnishes clinicians with a convenient yet robust tool for risk stratification and timely intervention in critical‑care settings.
			\begin{flushleft}
			\end{flushleft}
		\end{abstract}
		\newpage
		\tableofcontents 
		\newpage
		
		\begin{multicols}{2}
		\raggedcolumns  
		
		\vspace{-0.5em}   
		\section{Introduction}
		\setlength{\parskip}{0.3em}
		The precise and early diagnosis of mortality risk in hospitalized patients, particularly those in Intensive Care Units (ICUs), is a major challenge in modern medicine.  With ICU mortality rates ranging from 9.3\% to 26.2\% worldwide and respiratory illness mortality approaching one in every four patients, doctors encounter a significant cognitive load when monitoring numerous patients at once \cite{lee_mortality_2016}. This high-risk setting demands intelligent technologies capable of supplementing human expertise and facilitating early intervention in order to improve patient safety and care quality.
		
		Traditional severity grading methods such as APACHE, SAPS, and SOFA have made significant contributions but have well-documented drawbacks.  These systems are essentially static, collecting data from the first 24 hours after ICU admission and may not catch later changes in patient status.  Their moderate prediction accuracy (AUROC values ranging from 0.65 to 0.85 \cite{wang_crisp_2025}) highlights the need for more robust forecasting approaches.
		
		While machine learning approaches have showed promise, with some research obtaining AUROC values more than 0.85, comprehensive reviews demonstrate that when used generally, ML models often yield only marginal improvements over traditional scoring systems.  One important shortcoming of current techniques is their inability to properly harness the rich information buried in unstructured clinical text, which accounts for around 80\% of electronic health record (EHR) data yet is typically underutilized.
		
		The introduction of Natural Language Processing (NLP) and transformer models such as BERT \cite{devlin_bert_2019} opens up new possibilities for utilizing clinical narratives.  These notes include important contextual elements and clinical reasoning that structured data lacks, allowing for more nuanced and precise risk prediction.  However, prior techniques have not fully taken use of the hierarchical representations that these models can give.
		
		To overcome these problems, we present ALFIA (Adaptive Layer Fusion with Intelligent Attention), a unique deep learning architecture intended specifically for identifying early mortality risk in ICU patients.  ALFIA expands on a LoRA-adapted \cite{hu_lora_2021} BERT foundation model by introducing a novel adaptive layer fusion method that dynamically integrates multi-layer semantic information, capturing both fine-grained clinical details and high-level medical ideas.  The architecture uses token-level attention methods for semantic fusion, allowing for exact detection of the most relevant clinical text parts.
		
		We test ALFIA against our newly developed cw-24 (CriticalWindow-24) standard, which focuses on the essential early 24-hour timeframe for risk assessment.  Our results show that ALFIA outperforms cutting-edge tabular classifiers in AUPRC while retaining balanced precision-recall performance.  Furthermore, ALFIA embeddings can be effortlessly linked with gradient boosting methods (ALFIA-boost) and deep neural networks (ALFIA-nn) to increase performance.
		
		This work introduces a novel architecture that efficiently uses normal clinical text for early mortality prediction, giving doctors a simple yet powerful tool for risk stratification and prompt intervention in critical care situations. 
	
		\section{Methods and Materials}
		\setlength{\parskip}{0.3em}
		In this section, we will go over the benchmark design, dataset processing, model architecture, and training and evaluation methods.
		
		\subsection{Design and Implementation of the CriticalWindow-24 Benchmark}
		The CriticalWindow-24 (CW-24) benchmark was created using two large-scale critical care databases (\hyperref[fig:figure1]{Figure 1A}): MIMIC-IV (65,366 ICU admissions, 10.84\% death) and the eICU Collaborative Research Database (157,883 ICU admissions, 8.77\% mortality).  The benchmark was developed using four essential principles: temporal integrity, data leakage prevention, clinical relevance, and standardized recording.  Baseline demographics, clinical severity scores (APACHE III/IV, SAPS II, OASIS, LODS, MELD, SIRS), dynamic clinical assessments (Glasgow Coma Scale and SOFA scores), and hospital mortality outcome were all included, with all predictive variables restricted to a 24-hour window following ICU admission.
		
		To prevent data leakage, strict temporal limitations were enforced via automated validation checks, guaranteeing that no future information after the 24-hour cutoff was incorporated in predictive variables.  Within the forecast frame, dynamic assessments were averaged using statistical metrics such as maximum, minimum, first, last, mean, and standard deviation.  The benchmark omitted those features that were unsuitable for training (such as patient id). \hyperref[tab:table1]{Table 1} and \hyperref[tab:table2]{2} show baseline characteristics stratified by hospital mortality result for the MIMIC-IV and eICU datasets, respectively, with detailed variable specifications available in the supplementary materials.
		\end{multicols}
		
		\begin{longtable}{@{}p{4cm}p{3.5cm}p{3.5cm}p{3.5cm}@{}}
			\caption{{\small \textbf{Baseline Characteristics by Hospital Mortality Outcome}}} 
			\label{tab:table1} \\
			\toprule
			\textbf{Characteristic} & \textbf{Survived (N=58,280)} & \textbf{Died (N=7,086)} & \textbf{Overall (N=65,366)} \\
			\midrule
			\endfirsthead
			
			\caption[]{\textbf{Baseline Characteristics by Hospital Mortality Outcome (continued)}} \\
			\toprule
			\textbf{Characteristic} & \textbf{Survived (N=58,280)} & \textbf{Died (N=7,086)} & \textbf{Overall (N=65,366)} \\
			\midrule
			\endhead
			
			\midrule
			\multicolumn{4}{r}{\textit{Continued on next page}} \\
			\endfoot
			
			\bottomrule
			\endlastfoot
			
			\multicolumn{4}{l}{\textbf{Demographics \& Anthropometrics}} \\
			\midrule
			Admission Age (Mean ± SD) & 64.30 ± 17.17 & \textbf{71.08 ± 15.53} & 65.03 ± 17.13 \\
			Height (Mean ± SD) & 169.93 ± 10.61 & 168.22 ± 10.62 & 169.72 ± 10.63 \\
			Weight (Mean ± SD) & 82.05 ± 34.86 & 78.68 ± 24.32 & 81.69 ± 33.90 \\
			\midrule
			
			\multicolumn{4}{l}{\textbf{Clinical Severity Scores}} \\
			\midrule
			APACHE III (Mean ± SD) & 38.90 ± 17.16 & \textbf{65.53 ± 26.95} & 41.79 ± 20.24 \\
			SAPS II (Mean ± SD) & 32.90 ± 12.62 & \textbf{50.24 ± 16.24} & 34.78 ± 14.13 \\
			LODS (Mean ± SD) & 3.72 ± 2.56 & \textbf{7.18 ± 3.55} & 4.10 ± 2.89 \\
			MELD (Mean ± SD) & 12.50 ± 6.92 & \textbf{19.47 ± 10.06} & 13.26 ± 7.64 \\
			OASIS (Mean ± SD) & 29.58 ± 7.96 & \textbf{38.63 ± 9.06} & 30.57 ± 8.56 \\
			\midrule
			
			\multicolumn{4}{l}{\textbf{Dynamic Clinical Assessments - Glasgow Coma Scale}} \\
			\midrule
			GCS Maximum (Mean ± SD) & 14.88 ± 0.59 & 14.59 ± 1.50 & 14.85 ± 0.75 \\
			GCS Minimum (Mean ± SD) & 13.82 ± 2.48 & \textbf{12.76 ± 3.70} & 13.70 ± 2.66 \\
			GCS First (Mean ± SD) & 14.40 ± 1.96 & 14.04 ± 2.40 & 14.36 ± 2.02 \\
			GCS Last (Mean ± SD) & 14.65 ± 1.12 & \textbf{13.89 ± 2.72} & 14.57 ± 1.40 \\
			GCS Average (Mean ± SD) & 14.56 ± 1.00 & \textbf{14.01 ± 1.98} & 14.50 ± 1.16 \\
			GCS Std Dev (Mean ± SD) & 0.45 ± 1.01 & \textbf{0.83 ± 1.53} & 0.49 ± 1.08 \\
			\midrule
			
			\multicolumn{4}{l}{\textbf{Dynamic Clinical Assessments - SOFA Score}} \\
			\midrule
			SOFA Maximum (Mean ± SD) & 3.66 ± 2.83 & \textbf{6.92 ± 4.18} & 4.02 ± 3.18 \\
			SOFA Minimum (Mean ± SD) & 1.42 ± 1.91 & \textbf{2.71 ± 2.87} & 1.56 ± 2.08 \\
			SOFA First (Mean ± SD) & 1.47 ± 1.95 & \textbf{2.77 ± 2.92} & 1.61 ± 2.12 \\
			SOFA Last (Mean ± SD) & 3.56 ± 2.80 & \textbf{6.78 ± 4.16} & 3.91 ± 3.14 \\
			SOFA Average (Mean ± SD) & 3.05 ± 2.51 & \textbf{5.63 ± 3.57} & 3.33 ± 2.77 \\
			SOFA Std Dev (Mean ± SD) & 0.70 ± 0.63 & \textbf{1.28 ± 1.01} & 0.76 ± 0.70 \\
			\midrule
			
			\multicolumn{4}{l}{\textbf{Categorical Variables}} \\
			\midrule
			\textbf{Gender (n (\%))} & & & \\
			\quad Female & 25,402 (43.59\%) & 3,244 (45.78\%) & 28,646 (43.82\%) \\
			\quad Male & 32,878 (56.41\%) & 3,842 (54.22\%) & 36,720 (56.18\%) \\
			\midrule
			\textbf{Marital Status (n (\%))} & & & \\
			\quad Divorced & 4,138 (7.10\%) & 405 (5.72\%) & 4,543 (6.95\%) \\
			\quad Married & 26,509 (45.49\%) & 2,740 (38.67\%) & 29,249 (44.75\%) \\
			\quad Single & 15,944 (27.36\%) & 1,426 (20.12\%) & 17,370 (26.57\%) \\
			\quad Widowed & 6,479 (11.12\%) & \textbf{1,026 (14.48\%)} & 7,505 (11.48\%) \\
			\quad Unknown & 5,210 (8.94\%) & \textbf{1,489 (21.01\%)} & 6,699 (10.25\%) \\
			\midrule
			\textbf{Insurance (n (\%))} & & & \\
			\quad Medicaid & 8,502 (14.59\%) & 851 (12.01\%) & 9,353 (14.31\%) \\
			\quad Medicare & 29,947 (51.38\%) & \textbf{4,517 (63.75\%)} & 34,464 (52.72\%) \\
			\quad No Charge & 6 (0.01\%) & 1 (0.01\%) & 7 (0.01\%) \\
			\quad Other & 1,608 (2.76\%) & 125 (1.76\%) & 1,733 (2.65\%) \\
			\quad Private & 17,172 (29.46\%) & 1,275 (17.99\%) & 18,447 (28.22\%) \\
			\quad Unknown & 1,045 (1.79\%) & 317 (4.47\%) & 1,362 (2.08\%) \\
			\midrule
			\textbf{Smoker (n (\%))} & & & \\
			\quad No & 54,432 (93.40\%) & 6,742 (95.15\%) & 61,174 (93.59\%) \\
			\quad Yes & 3,848 (6.60\%) & 344 (4.85\%) & 4,192 (6.41\%) \\
			\midrule
			\textbf{Alcohol Abuse (n (\%))} & & & \\
			\quad No & 57,733 (99.06\%) & 7,040 (99.35\%) & 64,773 (99.09\%) \\
			\quad Yes & 547 (0.94\%) & 46 (0.65\%) & 593 (0.91\%) \\
			\midrule
			\textbf{SIRS Score (n (\%))} & & & \\
			\quad 0 & 1,418 (2.43\%) & 46 (0.65\%) & 1,464 (2.24\%) \\
			\quad 1 & 8,628 (14.80\%) & 377 (5.32\%) & 9,005 (13.78\%) \\
			\quad 2 & 19,723 (33.84\%) & 1,665 (23.50\%) & 21,388 (32.72\%) \\
			\quad 3 & 21,187 (36.35\%) & 3,024 (42.68\%) & 24,211 (37.04\%) \\
			\quad 4 & 7,324 (12.57\%) & \textbf{1,974 (27.86\%)} & 9,298 (14.22\%) \\
			
		\end{longtable}
		
		\begin{longtable}{@{}p{4cm}p{3.5cm}p{3.5cm}p{3.5cm}@{}}
			\caption{{\small \textbf{Baseline Characteristics by Hospital Mortality Outcome}}} 
			\label{tab:table2} \\
			\toprule
			\textbf{Characteristic} & \textbf{Survived (N=142,628)} & \textbf{Died (N=13,838)} & \textbf{Overall (N=157,883)} \\
			\midrule
			\endfirsthead
			
			\caption[]{\textbf{Baseline Characteristics by Hospital Mortality Outcome (continued)}} \\
			\toprule
			\textbf{Characteristic} & \textbf{Survived (N=142,628)} & \textbf{Died (N=13,838)} & \textbf{Overall (N=157,883)} \\
			\midrule
			\endhead
			
			\midrule
			\multicolumn{4}{r}{\textit{Continued on next page}} \\
			\endfoot
			
			\bottomrule
			\endlastfoot
			
			\multicolumn{4}{l}{\textbf{Demographics \& Anthropometrics}} \\
			\midrule
			Age (Mean ± SD) & 62.43 ± 17.25 & \textbf{69.76 ± 15.06} & 63.10 ± 17.20 \\
			Height (Mean ± SD) & 169.35 ± 13.74 & 168.42 ± 14.47 & 169.25 ± 13.87 \\
			Weight (Mean ± SD) & 84.19 ± 26.84 & 80.90 ± 28.18 & 83.89 ± 26.97 \\
			\midrule
			
			\multicolumn{4}{l}{\textbf{Clinical Severity Scores}} \\
			\midrule
			APACHE IV (Mean ± SD) & 51.02 ± 22.63 & \textbf{87.23 ± 33.34} & 54.21 ± 25.89 \\
			SAPS II (Mean ± SD) & 28.90 ± 13.05 & \textbf{49.58 ± 17.89} & 30.74 ± 14.76 \\
			OASIS (Mean ± SD) & 24.65 ± 8.91 & \textbf{35.76 ± 11.34} & 25.63 ± 9.68 \\
			\midrule
			
			\multicolumn{4}{l}{\textbf{Dynamic Clinical Assessments - Glasgow Coma Scale}} \\
			\midrule
			GCS Maximum (Mean ± SD) & 14.28 ± 1.88 & \textbf{11.03 ± 4.38} & 13.99 ± 2.39 \\
			GCS Minimum (Mean ± SD) & 12.72 ± 3.55 & \textbf{8.47 ± 4.73} & 12.35 ± 3.86 \\
			GCS First (Mean ± SD) & 13.28 ± 3.20 & \textbf{10.19 ± 4.75} & 13.01 ± 3.47 \\
			GCS Last (Mean ± SD) & 13.93 ± 2.33 & \textbf{9.40 ± 4.64} & 13.53 ± 2.90 \\
			GCS Average (Mean ± SD) & 13.63 ± 2.39 & \textbf{9.76 ± 4.35} & 13.29 ± 2.84 \\
			GCS Std Dev (Mean ± SD) & 0.72 ± 1.24 & \textbf{1.27 ± 1.57} & 0.77 ± 1.28 \\
			\midrule
			
			\multicolumn{4}{l}{\textbf{Dynamic Clinical Assessments - SOFA Score}} \\
			\midrule
			SOFA Maximum (Mean ± SD) & 4.58 ± 3.01 & \textbf{7.83 ± 3.95} & 4.86 ± 3.24 \\
			SOFA Minimum (Mean ± SD) & 1.84 ± 2.13 & \textbf{3.42 ± 2.91} & 1.98 ± 2.25 \\
			SOFA First (Mean ± SD) & 2.00 ± 2.25 & \textbf{3.56 ± 2.99} & 2.14 ± 2.37 \\
			SOFA Last (Mean ± SD) & 4.30 ± 2.95 & \textbf{7.55 ± 3.98} & 4.58 ± 3.19 \\
			SOFA Average (Mean ± SD) & 3.92 ± 2.74 & \textbf{6.67 ± 3.49} & 4.16 ± 2.92 \\
			SOFA Std Dev (Mean ± SD) & 0.86 ± 0.70 & \textbf{1.39 ± 1.03} & 0.90 ± 0.75 \\
			\midrule
			
			\multicolumn{4}{l}{\textbf{Categorical Variables}} \\
			\midrule
			\textbf{Region (n (\%))} & & & \\
			\quad Midwest & 49,371 (34.62\%) & 4,127 (29.82\%) & 54,232 (34.35\%) \\
			\quad Northeast & 10,104 (7.08\%) & \textbf{1,315 (9.50\%)} & 11,459 (7.26\%) \\
			\quad South & 44,411 (31.14\%) & \textbf{4,741 (34.26\%)} & 49,403 (31.29\%) \\
			\quad West & 29,530 (20.70\%) & 2,841 (20.53\%) & 32,676 (20.70\%) \\
			\quad Unknown & 9,212 (6.46\%) & 814 (5.88\%) & 10,113 (6.41\%) \\
			\midrule
			\textbf{Ethnicity (n (\%))} & & & \\
			\quad African American & 15,966 (11.19\%) & 1,423 (10.28\%) & 17,536 (11.11\%) \\
			\quad Asian & 2,389 (1.67\%) & 251 (1.81\%) & 2,677 (1.70\%) \\
			\quad Caucasian & 109,236 (76.59\%) & 10,724 (77.50\%) & 121,023 (76.65\%) \\
			\quad Hispanic & 5,452 (3.82\%) & 549 (3.97\%) & 6,031 (3.82\%) \\
			\quad Native American & 1,050 (0.74\%) & 94 (0.68\%) & 1,150 (0.73\%) \\
			\quad Other/Unknown & 6,776 (4.75\%) & 626 (4.52\%) & 7,504 (4.75\%) \\
			\quad Unknown & 1,759 (1.23\%) & 171 (1.24\%) & 1,962 (1.24\%) \\
			\midrule
			\textbf{Unit Type (n (\%))} & & & \\
			\quad CCU-CTICU & 12,099 (8.48\%) & 1,077 (7.78\%) & 13,218 (8.37\%) \\
			\quad CSICU & 5,264 (3.69\%) & 331 (2.39\%) & 5,633 (3.57\%) \\
			\quad CTICU & 4,754 (3.33\%) & 294 (2.12\%) & 5,087 (3.22\%) \\
			\quad Cardiac ICU & 9,932 (6.96\%) & \textbf{1,131 (8.17\%)} & 11,176 (7.08\%) \\
			\quad MICU & 11,648 (8.17\%) & \textbf{1,607 (11.61\%)} & 13,365 (8.47\%) \\
			\quad Med-Surg ICU & 79,199 (55.53\%) & 7,629 (55.13\%) & 87,717 (55.56\%) \\
			\quad Neuro ICU & 10,775 (7.55\%) & 956 (6.91\%) & 11,863 (7.51\%) \\
			\quad SICU & 8,957 (6.28\%) & 813 (5.88\%) & 9,824 (6.22\%) \\
			\midrule
			\textbf{Gender (n (\%))} & & & \\
			\quad Female & 65,530 (45.94\%) & 6,446 (46.58\%) & 72,670 (46.03\%) \\
			\quad Male & 77,052 (54.02\%) & 7,376 (53.30\%) & 85,131 (53.92\%) \\
			\quad Unknown & 46 (0.03\%) & 16 (0.12\%) & 82 (0.05\%) \\
			\midrule
			\textbf{Smoker (n (\%))} & & & \\
			\quad No & 12,022 (8.43\%) & 1,136 (8.21\%) & 13,386 (8.48\%) \\
			\quad Yes & 14,331 (10.05\%) & 1,263 (9.13\%) & 15,889 (10.06\%) \\
			\quad Unknown & 116,275 (81.52\%) & 11,439 (82.66\%) & 128,608 (81.46\%) \\
			\midrule
			\textbf{Alcohol Abuse (n (\%))} & & & \\
			\quad No & 139,649 (97.91\%) & 13,612 (98.37\%) & 154,643 (97.95\%) \\
			\quad Yes & 2,979 (2.09\%) & 226 (1.63\%) & 3,240 (2.05\%) \\
			\midrule
			\textbf{Drug Abuse (n (\%))} & & & \\
			\quad No & 142,453 (99.88\%) & 13,833 (99.96\%) & 157,701 (99.88\%) \\
			\quad Yes & 175 (0.12\%) & 5 (0.04\%) & 182 (0.12\%) \\
			\midrule
			\textbf{Obesity (n (\%))} & & & \\
			\quad No & 141,141 (98.96\%) & 13,664 (98.74\%) & 156,203 (98.94\%) \\
			\quad Yes & 1,487 (1.04\%) & 174 (1.26\%) & 1,680 (1.06\%) \\
			
		\end{longtable}
		
		\begin{multicols}{2}
		\subsection{Dataset Processing and Text Encoding}
\begin{figure}[htbp]           
		\centering
		\includegraphics[width=1\linewidth]{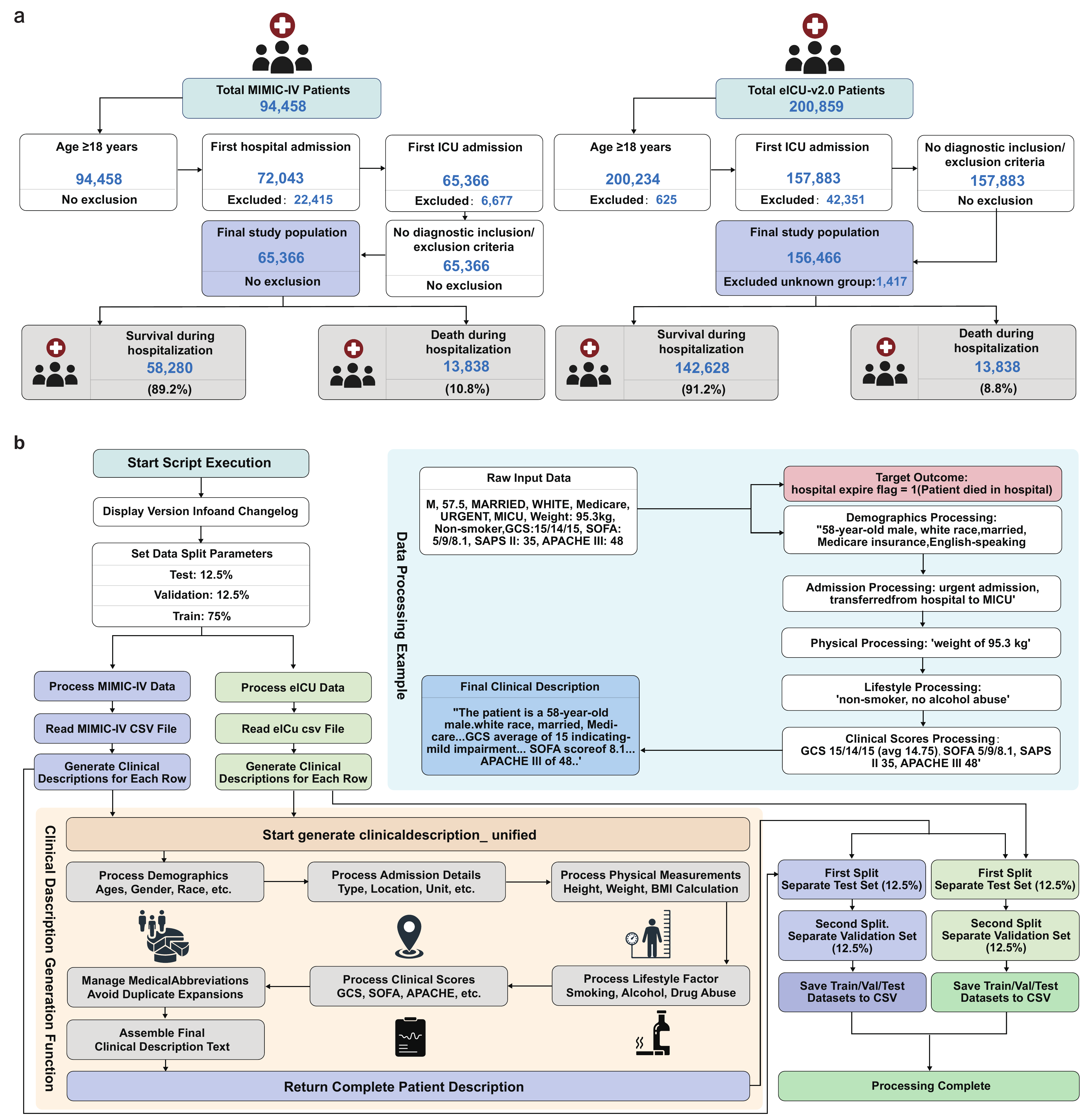}
		\label{fig:figure1}         
		\end{figure}

		\setlength{\parskip}{0.3em}
		We processed the datasets and encoded the text using a uniform pipeline script. To encode tabular data into fluent clinical descriptive text, we used a consistent process across both the MIMIC-IV \cite{johnson_mimic-iv_2023} and eICU \cite{pollard_eicu_2018} datasets (\hyperref[fig:figure1]{Figure 1B}). The pipeline processes demographics (age, gender, race, etc.), admission details (type, location, unit, etc.), physical measurements (height, weight, BMI calculation), lifestyle factors (smoking, alcohol, drug abuse), clinical scores (GCS, SOFA, APACHE, etc.), and medical abbreviations (avoiding duplicate expansions). Finally, the datasets were divided equally into 75-12.5-12.5 train-validation-test formats for all downstream model training (including ALFIA and other comparable models).
		
		\subsection{The ALFIA Architecture}
\begin{figure}[htbp]           
		\centering
		\includegraphics[width=1\linewidth]{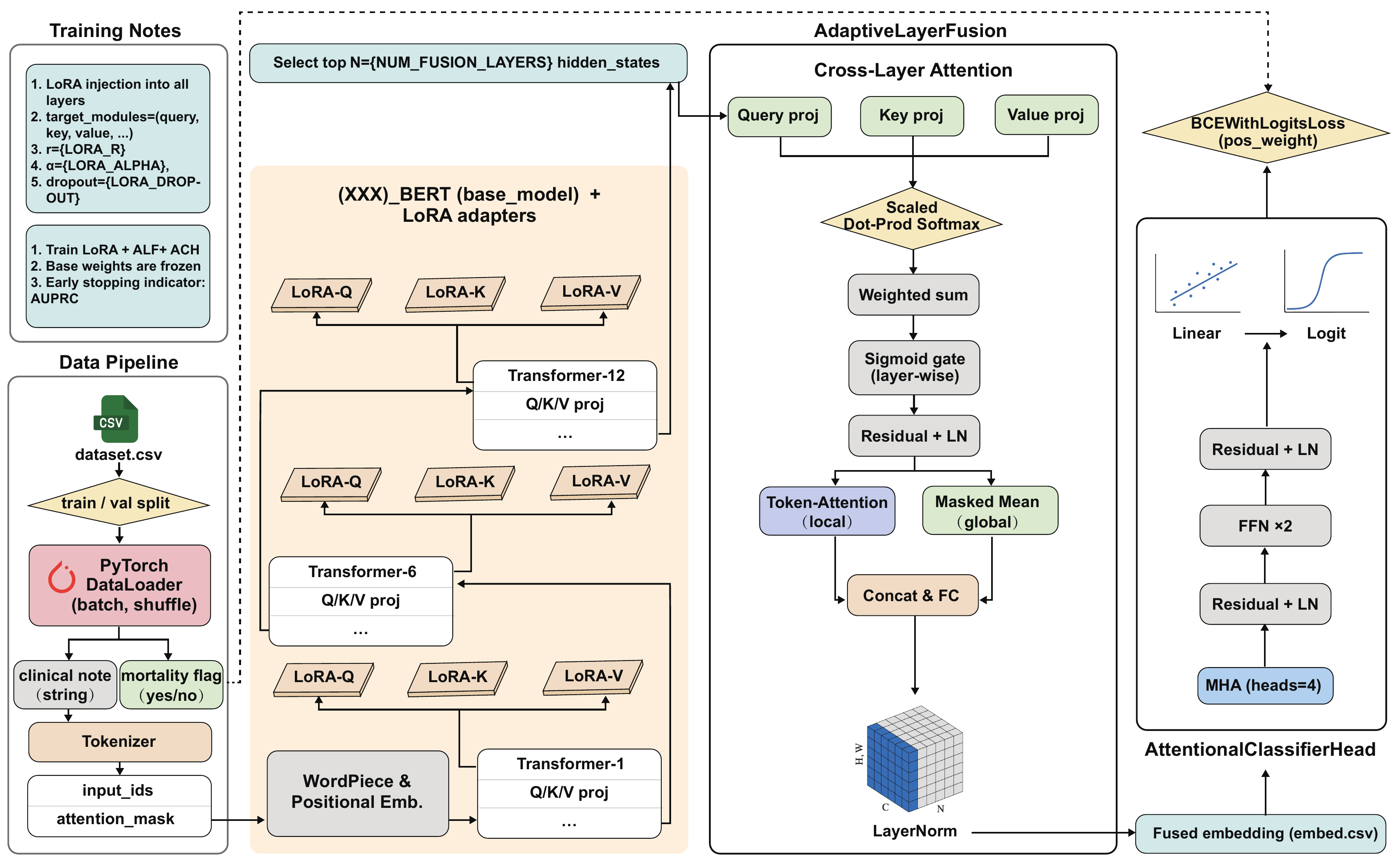}
		\label{fig:figure2}         
		\end{figure}

		\setlength{\parskip}{0.3em}
		The proposed model architecture, ALFIA (Adaptive Layer Fusion Integrated Architecture), is intended to efficiently use hierarchical features from pre-trained transformer models for text classification tasks (\hyperref[fig:figure2]{Figure 2}). It has three primary sequential components: a Base Transformer Model, an Adaptive Layer Fusion (ALF) module, and an Attentional Classifier Head. Low-Rank Adaptation (LoRA) is an option for efficient fine-tuning.
		
		\subsubsection{Base Transformer Model}
		\setlength{\parskip}{0.3em}
		The foundation of the ALFIA architecture is a pre-trained Base Transformer \cite{vaswani_attention_2023} Model, such as BERT \cite{devlin_bert_2019}, RoBERTa \cite{liu_roberta_2019}, BioBERT \cite{lee_biobert_2020}, or other similar encoder-based architectures. This model serves as the primary feature extractor. Given an input sequence of tokens $X = \{x_1, x_2, \dots, x_T\}$, where $T$ is the sequence length, the Base Transformer Model outputs a series of hidden state sequences from its $L$ layers:
		\begin{equation} \label{eq:base_hidden_states}
			H^{(0)}, H^{(1)}, \dots, H^{(L)}
		\end{equation}
		Here, $H^{(0)}$ represents the initial token and positional embeddings. For each layer $l \in [1, L]$, $H^{(l)} = \{h_1^{(l)}, h_2^{(l)}, \dots, h_T^{(l)}\}$ is the sequence of hidden states, where each $h_t^{(l)} \in \mathbb{R}^{d_{\text{model}}}$ is the hidden state for token $t$ at layer $l$, and $d_{\text{model}}$ is the dimensionality of the hidden states.
		
		\end{multicols}
		
		\captionof{figure}{\footnotesize \textbf{Figure 1. Patient Selection and Data Processing Pipeline.} \textbf{(a)} Systematic patient selection strategy from MIMIC-IV and eICU 2.0 databases: inclusion criteria include ICU admission, age $\geq$18 years, and first-time hospitalization or ICU admission, with no diagnostic exclusions applied to obtain the final sample for downstream processing. \textbf{(b)} Clinical description encoding workflow transforming tabular patient data into coherent textual descriptions, sequentially processing demographics, admission details, basic physiological parameters, medical history, and clinical scoring systems, culminating in dataset partitioning into training, validation, and test sets.}
		
		\begin{multicols}{2}
			
		\subsubsection{Adaptive Layer Fusion (ALF) Module}
		\setlength{\parskip}{0.3em}
		The ALF module is a critical component designed to dynamically integrate information from multiple layers of the Base Transformer Model. It takes as input the hidden states from the top $N_f$ layers, i.e., $\{H^{(L-N_f+1)}, \dots, H^{(L)}\}$. The objective is to learn a weighted combination of these layer representations, allowing the model to capture a richer set of features spanning different levels of abstraction.
		
		The ALF module operates in several stages:
		\paragraph{Layer Weight Computation:} This stage determines the importance (weight) $\lambda_j$ for each of the $N_f$ selected layers.
		First, an input layer summary $s_j$ for each layer $j$ (from the $N_f$ layers) is computed via attention-mask-weighted average pooling of its token hidden states $h_t^{(j)}$:
		\begin{equation} \label{eq:alf_layer_summary}
			s_j = \frac{\sum_{t=1}^{T} A_t \cdot h_t^{(j)}}{\sum_{t=1}^{T} A_t}
		\end{equation}
		where $A_t$ is the attention mask ($A_t=1$ for real tokens, $0$ for padding). These $N_f$ summary vectors $\{s_1, \dots, s_{N_f}\}$, each in $\mathbb{R}^{d_{\text{model}}}$, are stacked into $S_{\text{stack}} \in \mathbb{R}^{N_f \times d_{\text{model}}}$.
		Next, an inter-layer attention mechanism computes layer contributions. A global query vector $q_{\text{global}} = \text{mean}(s_1, \dots, s_{N_f})$ is formed. This query and $S_{\text{stack}}$ are projected into Query ($Q_{\text{proj}}$), Key ($K_{\text{proj}}$), and Value ($V_{\text{proj}}$) spaces:
		\begin{align}
			Q_{\text{proj}} &= q_{\text{global}}W_Q \in \mathbb{R}^{d_k \cdot n_h} \label{eq:alf_q_proj} \\
			K_{\text{proj}} &= S_{\text{stack}}W_K \in \mathbb{R}^{N_f \times (d_k \cdot n_h)} \label{eq:alf_k_proj} \\
			V_{\text{proj}} &= S_{\text{stack}}W_V \in \mathbb{R}^{N_f \times (d_v \cdot n_h)} \label{eq:alf_v_proj}
		\end{align}
		where $W_Q, W_K, W_V$ are trainable weight matrices, $n_h$ is the number of attention heads, and $d_k, d_v$ are head dimensions. Attention scores are calculated as:
		{\footnotesize
			\begin{equation} \label{eq:alf_attn_scores}
				\begin{split}
					\text{scores} ={}& \text{softmax}\left(\frac{Q_{\text{proj}}K_{\text{proj}}^T}{\sqrt{d_k}}\right) \\
					& \in \mathbb{R}^{1 \times N_f} \quad (\text{per head})
				\end{split}
			\end{equation}
		}
		The output context $c_{\text{layer\_attn}} = \text{scores} \cdot V_{\text{proj}}$ is projected to $c'_{\text{layer\_attn}} = c_{\text{layer\_attn}}W_O \in \mathbb{R}^{d_{\text{model}}}$.
		Optionally, if layer gating is enabled, these are passed through a linear layer and sigmoid:
		\begin{equation} \label{eq:alf_lambda_weights}
			\lambda = \text{sigmoid}(c'_{\text{layer\_attn}}W_{\text{gate}} + b_{\text{gate}}) \in \mathbb{R}^{N_f}
		\end{equation}
		Otherwise, normalized attention scores are used as $\lambda$.
		
		\paragraph{Weighted Combination of Layer Hidden States:} The full hidden state sequences from the selected $N_f$ layers are combined using the learned layer weights $\lambda_j$. For each token position $t$:
		\begin{equation} \label{eq:alf_fused_raw}
			h_t^{\text{fused\_raw}} = \sum_{j=1}^{N_f} \lambda_j \cdot h_t^{(L-N_f+j)}
		\end{equation}
		This results in a sequence $H^{\text{fused\_raw}} \in \mathbb{R}^{T \times d_{\text{model}}}$.
		
		\paragraph{Post-Fusion Processing:} The raw fused states $H^{\text{fused\_raw}}$ undergo further transformations, including Layer Interaction, Content Projection, and an Enhancement step:
		{\footnotesize
			\begin{equation} \label{eq:alf_enhanced_states}
				\begin{split}
					H^{\text{enhanced}} ={}& \text{LayerNorm}_1(H^{\text{fused\_raw}} \\
					& + H^{\text{projected}} + H^{\text{interaction}})
				\end{split}
			\end{equation}
		}
		These operations typically involve sequences of Linear, Layer Normalization, GELU, and Dropout layers.
		
		\paragraph{Token-Level Attention for Local Context:} The enhanced states $H^{\text{enhanced}}$ are processed by a token-level attention mechanism. Token scores $\text{score}_t$ are computed for each token $h_t^{\text{enhanced}}$, masked, and softmaxed to yield token weights $\beta_t$:
		\begin{equation} \label{eq:alf_beta_weights}
			\beta = \text{softmax}(\text{scores}_{\text{masked}})
		\end{equation}
		The local context $c_{\text{local}}$ is then a weighted sum:
		{\footnotesize
			\begin{equation} \label{eq:alf_local_context}
				\begin{split}
					c_{\text{local}} ={}& \sum_{t=1}^{T} \beta_t \cdot h_t^{\text{enhanced}} \\
					& \in \mathbb{R}^{d_{\text{model}}}
				\end{split}
			\end{equation}
		}
		
		\paragraph{Global Context Extraction:} A global context vector $c_{\text{global}}$ is derived from $H^{\text{enhanced}}$ via attention-mask-weighted average pooling, followed by a global context processing layer:
		{\footnotesize
			\begin{align}
				c_{\text{global\_pooled}} &= \frac{\sum_{t=1}^{T} A_t \cdot h_t^{\text{enhanced}}}{\sum_{t=1}^{T} A_t} \label{eq:alf_global_pooled} \\
				\begin{split} \label{eq:alf_global_context}
					c_{\text{global}} ={}& \text{GlobalContextLayer}(c_{\text{global\_pooled}}) \\
					& \in \mathbb{R}^{d_{\text{model}}}
				\end{split}
			\end{align}
		}
		
		\paragraph{Final Context Fusion and Output:} The local and global context vectors are concatenated, $[c_{\text{local}}; c_{\text{global}}] \in \mathbb{R}^{2 \cdot d_{\text{model}}}$, and then fused by a Context Fusion layer to produce $c_{\text{fused}} \in \mathbb{R}^{d_{\text{model}}}$. An output projection and LayerNorm yield the final ALF output vector $H_{\text{ALF\_out}}$:
		{\footnotesize
			\begin{equation} \label{eq:alf_out_final}
				\begin{split}
					H_{\text{ALF\_out}} ={}& \text{LayerNorm}_2(c_{\text{fused}} \\
					& + \text{OutputProjection}(c_{\text{fused}})) \\
					& \in \mathbb{R}^{d_{\text{model}}}
				\end{split}
			\end{equation}
		}
		This vector $H_{\text{ALF\_out}}$ serves as the enriched representation for the subsequent classifier.
		
		\subsubsection{Attentional Classifier Head (ACH) Module}
		The fused embedding $H_{\text{ALF\_out}}$ (denoted as $Z$ for simplicity) from the ALF module is processed by the Attentional Classifier Head.
		First, $Z$ is unsqueezed to $Z' \in \mathbb{R}^{1 \times d_{\text{model}}}$ to be compatible with attention mechanisms expecting sequence input. It then undergoes multi-head self-attention:
		{\footnotesize
			\begin{equation} \label{eq:classifier_attn_out}
				\begin{split}
					\text{AttnOut} ={}& \text{MultiheadAttention}(Q=Z', \\
					& \quad K=Z', V=Z') 
				\end{split}
			\end{equation}
		}
		This allows features within the fused representation $Z$ to interact and be re-weighted. Following this, standard transformer block operations, including Layer Normalization, residual connections, and a Feed-Forward Network (FFN), are applied:
		{\footnotesize
			\begin{align}
				\begin{split}\label{eq:classifier_z_norm1}
					Z'_{\text{norm1}} ={}& \text{LayerNorm}_1(Z' \\
					& + \text{Dropout}(\text{AttnOut}))
				\end{split} \\
				Z'_{\text{ffn}} &= \text{FFN}(Z'_{\text{norm1}}) \label{eq:classifier_z_ffn} \\
				\begin{split}\label{eq:classifier_z_norm2}
					Z'_{\text{norm2}} ={}& \text{LayerNorm}_2(Z'_{\text{norm1}} \\
					& + \text{Dropout}(Z'_{\text{ffn}}))
				\end{split}
			\end{align}
		}
		The processed vector $Z'_{\text{norm2}}$ (squeezed back to $\mathbb{R}^{d_{\text{model}}}$) is then passed to an output linear layer to produce the logits for classification:
		\begin{equation} \label{eq:classifier_logits}
			\text{logits} = Z'_{\text{norm2}}W_{\text{out}} + b_{\text{out}}
		\end{equation}
		where $W_{\text{out}}$ and $b_{\text{out}}$ are the weight matrix and bias of the output layer, respectively. The logits are typically passed through a softmax function to obtain class probabilities.
		
		\subsubsection{Low-Rank Adaptation (LoRA)}
		\setlength{\parskip}{0.3em}
		To facilitate efficient fine-tuning, especially for large pre-trained models, Low-Rank Adaptation (LoRA) can be optionally integrated. When LoRA is enabled, for a pre-trained weight matrix $W_0 \in \mathbb{R}^{d_1 \times d_2}$ within the Base Transformer Model (e.g., in attention or feed-forward layers), its update $\Delta W$ is constrained to be of low rank. Specifically, $W_0$ is kept frozen, and two trainable low-rank matrices, $A \in \mathbb{R}^{d_1 \times r}$ and $B \in \mathbb{R}^{r \times d_2}$, are introduced, where the rank $r \ll \min(d_1, d_2)$. The forward pass for a layer is modified from $h = W_0x$ to:
		\begin{equation} \label{eq:lora_forward_pass}
			h = W_0x + BAx
		\end{equation}
		During fine-tuning, only the parameters of matrices $A$ and $B$ are updated, significantly reducing the number of trainable parameters compared to full fine-tuning. This approach is based on the hypothesis that the adaptation of pre-trained models to new tasks often occurs in a low-rank subspace of the weight parameter space.
		\end{multicols}
		\begin{algorithm}[H]  
			\caption{Model Architecture}
			\label{alg:simplified_model_architecture}
			\small  
			\begin{algorithmic}[1]
				\Require Raw text sequence $T$, Attention mask $M_{attn}$
				\Ensure Predicted class probabilities $P_{class}$
				
				\Statex \textit{// Initialize Components}
				\State BaseLM $\gets$ Pre-trained Transformer (e.g., BioBERT), optionally with LoRA
				\State FusionModule $\gets$ AdaptiveLayerFusion($H_{size}, N_{fuse}, \dots$)
				\State ClassifierHead $\gets$ AttentionalClassifierHead($H_{size}, N_{classes}, \dots$)
				
				\Statex \textit{// Forward Pass through Base Language Model}
				\State HiddenStates$_{all} \gets \text{BaseLM.forward}(T, M_{attn})$ \Comment{Get all layer hidden states}
				\State HiddenStates$_{fuse} \gets \text{SelectTopLayers}(\text{HiddenStates}_{all}, N_{fuse})$
				
				\Statex \textit{// Adaptive Layer Fusion}
				\Function{AdaptiveLayerFusion.process}{HS$_{fuse}$, $M_{attn}$}
				\State LayerRepresentations $\gets \text{AveragePoolPerLayer}(\text{HS}_{fuse}, M_{attn})$
				\State GateWeights $\gets \text{CalculateLayerAttentionWeights}(\text{LayerRepresentations})$ \Comment{Via self-attention and optional gating}
				\State WeightedStates $\gets \sum (\text{GateWeights} \times \text{HS}_{fuse})$ \Comment{Element-wise product and sum}
				\State EnhancedStates $\gets \text{ProcessWeightedStates}(\text{WeightedStates})$ \Comment{Includes interaction, projection, LayerNorm}
				\State LocalContext $\gets \text{TokenAttentionPool}(\text{EnhancedStates}, M_{attn})$
				\State GlobalContext $\gets \text{MaskedAveragePool}(\text{EnhancedStates}, M_{attn})$
				\State CombinedContext $\gets \text{FuseContexts}(\text{LocalContext}, \text{GlobalContext})$ \Comment{e.g., Concat + Linear + Norm}
				\State \Return CombinedContext, GateWeights
				\EndFunction
				\State FusedEmbed, LayerWeights $\gets$ FusionModule.process(HiddenStates$_{fuse}$, $M_{attn}$)
				
				\Statex \textit{// Attentional Classifier Head}
				\Function{AttentionalClassifierHead.process}{FusedEmbed}
				\State $X \gets \text{SelfAttentionBlock}(\text{FusedEmbed.unsqueeze}(1))$ \Comment{MHA, Add \& Norm, FFN, Add \& Norm}
				\State Logits $\gets \text{OutputLinearLayer}(X.\text{squeeze}(1))$
				\State \Return Logits
				\EndFunction
				\State Logits $\gets$ ClassifierHead.process(FusedEmbed)
				\State $P_{class} \gets \text{Sigmoid}(\text{Logits})$
			\end{algorithmic}
		\end{algorithm}
		
		\begin{algorithm}
			\caption{Inference Pipeline}
			\label{alg:simplified_inference_pipeline}
			\small
			\begin{algorithmic}[1]
				\Require List of texts $S_{texts}$, Model configurations (LM name, paths to weights, $N_{fuse}$, $L_{max}$), Batch size $B_{size}$
				\Ensure DataFrame of fused text embeddings $DF_{embed}$
				
				\Statex \textit{// 1. Initialization}
				\State BaseLM, FusionModule $\gets$ LoadModelsAndWeights(Model configurations)
				\State BaseLM.eval(), FusionModule.eval()
				\State Tokenizer $\gets$ LoadPretrainedTokenizer(LM name)
				\State AllEmbeddings $\gets []$, AllLayerWeights $\gets []$
				
				\Statex \textit{// 2. Process Texts in Batches}
				\For{each batch of texts in $S_{texts}$}
				\State EncodedInput, $M_{attn} \gets$ TokenizeBatch(BatchTexts, Tokenizer, $L_{max}$)
				\State \textbf{with} torch.no\_grad():
				\State HiddenStates$_{all} \gets \text{BaseLM.forward}(\text{EncodedInput['input\_ids']}, M_{attn})$
				\State HiddenStates$_{select} \gets \text{SelectTopLayers}(\text{HiddenStates}_{all}, N_{fuse})$
				\If{$N_{fuse} > 0$}
				\State FusedEmbeds$_{batch}$, Weights$_{batch} \gets$ FusionModule(HiddenStates$_{select}$, $M_{attn}$)
				\Else
				\State FusedEmbeds$_{batch} \gets \text{ZeroEmbeddingsForBatch}()$
				\State Weights$_{batch} \gets \text{UniformWeightsForBatch}()$
				\EndIf
				\State Append FusedEmbeds$_{batch}$ to AllEmbeddings
				\State Append Weights$_{batch}$ to AllLayerWeights
				\EndFor
				
				\Statex \textit{// 3. Finalize Output}
				\State EmbeddingsArray $\gets$ VStack(AllEmbeddings)
				\State LayerWeightsArray $\gets$ VStack(AllLayerWeights)
				\State NormalizedEmbeddings $\gets$ Normalize(EmbeddingsArray)
				\State $DF_{embed} \gets$ CreateDataFrame(NormalizedEmbeddings, LayerWeightsArray)
			\end{algorithmic}
		\end{algorithm}
		
		\begin{multicols}{2}
		\subsection{Training and Evaluation of ALFIA and Derivative Models, as well as Comparative Baseline Architectures}
		\setlength{\parskip}{0.3em}
		In all evaluation we use cw-24 MIMIC-IV subset as main benchmark and eicu subset as external validation. We conducted training and evaluation of ALFIA and its derivative models, as well as comparison baseline designs, through a set of unified pipeline scripts. All scripts were assigned the same random seed of 42 to ensure reproducibility. For ALFIA, we designed training scripts capable of interfacing with encoded text inputs, leveraging validation set AUPRC for evaluation and early termination (default patience of 5 epochs). We integrated real-time monitoring of AUPRC and AUROC measurements for each epoch, with automatic storage of the best-performing LoRA Adapter, ALF, and ACH states. Final test set evaluation involved full computation of categorization metrics and 95\% confidence intervals. For expert evaluation, we subset 100 samples randomly, expert should evaluate each sample and their confidence in each evaluation.
		
		For comparative baseline models, we employed the AutoGluon \cite{erickson_autogluon-tabular_2020} framework for unified training across a diverse ensemble of algorithms, including gradient boosting decision trees (LightGBM \cite{ke_lightgbm_2017}, XGBoost, CatBoost \cite{pollard_eicu_2018}), multi-layer neural networks (implemented using PyTorch and FastAI), traditional machine learning models (k-nearest neighbors, random forest), and emerging transformer-based architectures (TabPFN \cite{hollmann_accurate_2025}, FT-Transformers). All compared models use tabular data (encoded as text) as input, and utilized AUPRC as the primary training metric, with test set performance evaluated using identical classification metrics and 95\% confidence interval estimates. 
		
		For ALFIA derivative models (ALFIA-boost, ALFIA-nn), we exploited the embeddings created by ALFIA's ALF module alongside original features as inputs to the AutoGluon model ensemble, serving as alternatives to the ALFIA ACH for prediction tasks. Classification metrics and 95\% confidence intervals were obtained for all derivative model predictions.
		
		\subsection{Hardware and System Configuration}
		\setlength{\parskip}{0.3em}
		The experiments were carried out on cloud-based computing nodes with typical configurations. Each node included an NVIDIA RTX 4090 GPU with 24 GB of video RAM, a 16-core Intel Xeon(R) Platinum 8352V processor, and 120 GB of system memory. The nodes run Ubuntu 22.04 LTS, with GPU driver version 535.129.03 and CUDA support up to 12.6. All independent models were trained on a single GPU to maintain consistency in resource allocation and performance evaluation.
		
		The Mamba package manager was used to manage the software environment, and Python 3.11.12 served as the base interpreter. Key packages include pandas 2.2.3 for data manipulation, numpy 1.26.4 for numerical computing, matplotlib 3.10.1 and seaborn 0.12.2 for data visualization, and scikit-learn. 1.5.2 for machine learning utilities, torch 2.6.0 for deep learning framework support, transformers 4.49.0 for pre-trained model implementations, peft 0.15.2 for parameter-efficient fine-tuning, tqdm 4.67.1 for progress tracking, and autogluon 1.3.1 for automated batch machine learning. Portions of the statistical analysis and figure preparation were performed using GraphPad Prism version 10.4.2 software.
		
		\section{Results}
		\subsection{Training Dynamics Demonstrate ALFIA Convergence in Textual Data Learning}
\begin{figure}[htbp] 
			\centering
			\includegraphics[width=1\linewidth]{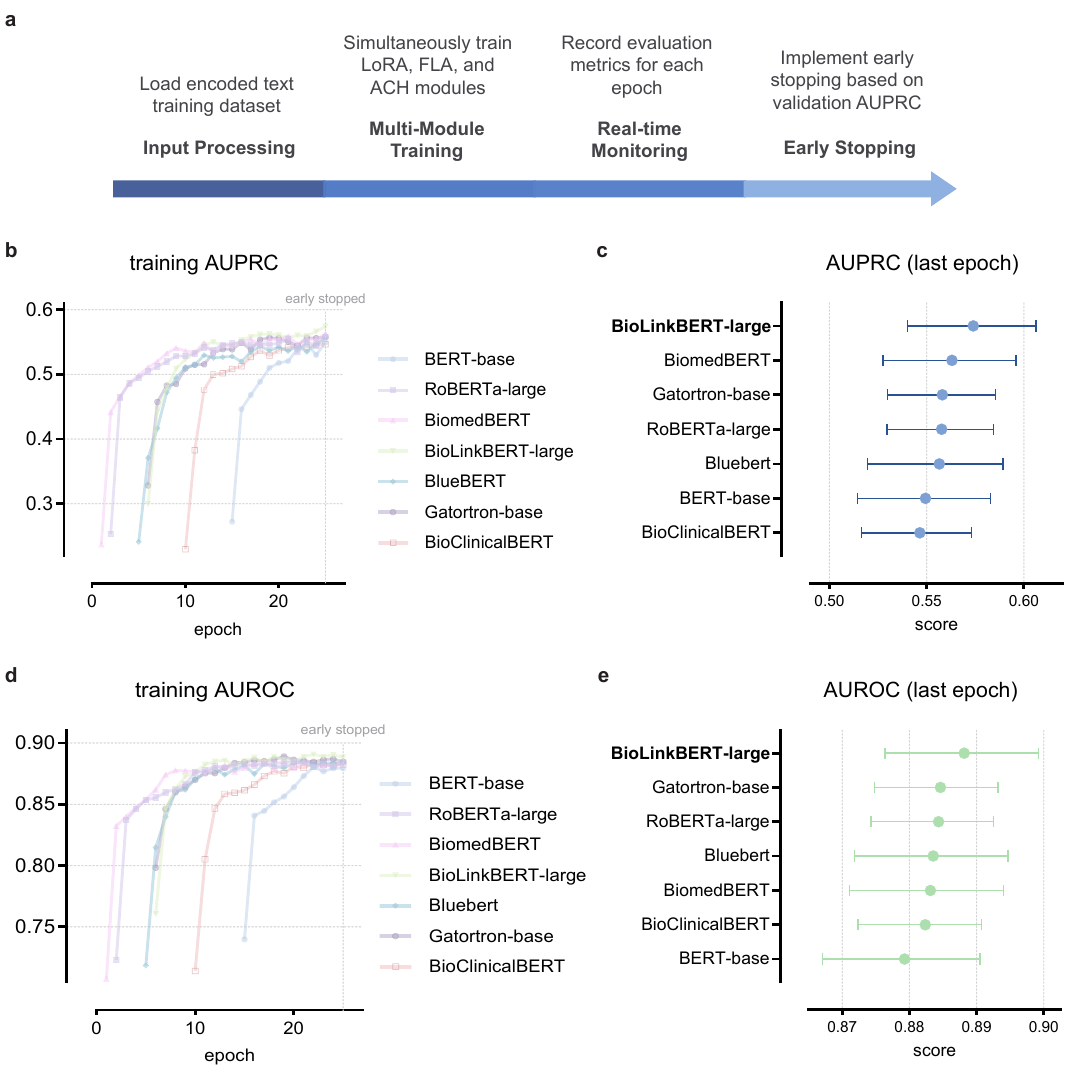} 
			\caption{\footnotesize \textbf{Figure 3. Training pipeline of ALFIA and performance comparison of different pre-trained language models.} \textbf{(a)} Overview of the training pipeline, including input processing, multi-module training with simultaneous optimization of LoRA, FLA, and ACH modules, real-time monitoring of evaluation metrics, and early stopping implementation based on validation AUPRC. \textbf{(b)} Training AUPRC curves across epochs for different pre-trained models. All models implement early stopping when validation performance plateaus. \textbf{(c)} Final AUPRC scores (last epoch) with 95\% confidence intervals. \textbf{(d)} Training AUROC curves across epochs for different pre-trained models.  \textbf{(e)} Final AUROC scores (last epoch) with 95\% confidence intervals.}
			\label{fig:figure3} 
		\end{figure}

		\setlength{\parskip}{0.3em}
		We conducted a systematic analysis of convergence performance across multiple pre-trained language models on the clinical mortality prediction task using various backbone models (\hyperref[fig:figure3]{Figure 3}). All models were trained with consistent hyperparameters (fusion layers: 4 layers; LoRA parameters: r=16, alpha=16, dropout=0.05,target-modules="query,key,value,output.dense"). The training process employed a synchronous multi-module training strategy, incorporating joint optimization of LoRA, FLA, and ACH modules, with early stopping implemented based on validation AUPRC to prevent overfitting (\hyperref[fig:figure3]{Figure 3a}).
		
		The training dynamics graphs show that all models had positive convergence characteristics throughout the training period.  In terms of the AUPRC measure, most models converged and stopped within 15-25 epochs.  BioLinkBERT-large outperformed the others, not only converging rapidly but also sustaining consistently high performance levels during training (\hyperref[fig:figure3]{Figure 3b}).  RoBERTa-large and BiomedBERT both have steady training trajectories with smooth climbing convergence curves.  In contrast, BERT-base and BioClinicalBERT demonstrated slower convergence rates, need more training epochs to reach stable states.
		
		Visualization of AUROC (\hyperref[fig:figure3]{Figure 3d}) measures throughout epochs revealed that all models had similar convergence trends, with the majority obtaining peak performance around 15 epochs.  Notably, models that were pre-trained on biomedical domains (such as BioLinkBERT-large, BiomedBERT, and Gatortron-base) performed better at the start of training, indicating that domain-specific pre-training does improve model performance on medical text understanding tasks.
		
		Final performance evaluation (\hyperref[fig:figure3]{Figure 3c}, \hyperref[fig:figure3]{3e}) demonstrated that BioLinkBERT-large outperformed both important parameters, with AUPRC averaging above 0.58 and AUROC reaching 0.89.  Gatortron-base and BiomedBERT followed closely behind, scoring around 0.57 and 0.56 on AUPRC, respectively.  These findings verify the success of our multi-module training technique while also highlighting the benefits of domain-adaptive pre-trained models in clinical prediction tasks.  This also demonstrates the effect of backbone models on ultimate performance.  All subsequent ALFIA backbone models will train with the best-performing BioLinkBERT.  The constant convergence of training trajectories, as well as the large improvement in final performance, suggest that the proposed method can effectively acquire valuable representations from clinical textual data, laying the groundwork for accurate mortality prediction.
		
		\end{multicols}
		
		\vspace{-10pt}
		\captionof{figure}{\footnotesize \textbf{Figure 2. The ALFIA Architecture for Clinical Mortality Prediction.} The framework consists of four main components: \textbf{(1) Data Pipeline}: Clinical notes are processed through tokenization and split into training/validation sets with mortality flags; \textbf{(2) Base Model with LoRA Adaptation}: A BERT-based transformer with Low-Rank Adaptation (LoRA) modules injected into query, key, and value projections across all layers, where base model weights remain frozen during training; \textbf{(3) AdaptiveLayerFusion Module}: Selects top-N hidden states from different transformer layers and applies cross-layer attention mechanism with query/key/value projections, followed by scaled dot-product attention, sigmoid gating, and residual connections. The fused representations undergo both local token-level attention and global masked mean pooling before concatenation; \textbf{(4) AttentionalClassifierHead}: Processes the fused embeddings through multi-head attention (4 heads), feed-forward networks with residual connections and layer normalization, culminating in a linear layer with logit output and BCEWithLogitsLoss for binary mortality prediction. Training employs LoRA with early stopping based on AUPRC metrics.}
		
		\begin{multicols}{2}
		
		\subsection{Superior Performance of ALFIA and ALFIA-boost/nn over Existing Methods}
		\setlength{\parskip}{0.3em}
		In real-world clinical and decision-making settings, mortality outcomes show a considerable class imbalance.   Hospital mortality makes up approximately 10\% of the total sample size in both the MIMIC and eICU datasets. Given that AUROC overestimates performance on imbalanced datasets, we use AUPRC (Area Under the Precision-Recall Curve) as our primary comparison and assessment metric.   We use threshold search methods on trained models to identify the best F1 and F2 scores that balance recall and precision.
		
\begin{figure}[htbp] 
			\centering
			\includegraphics[width=1\linewidth]{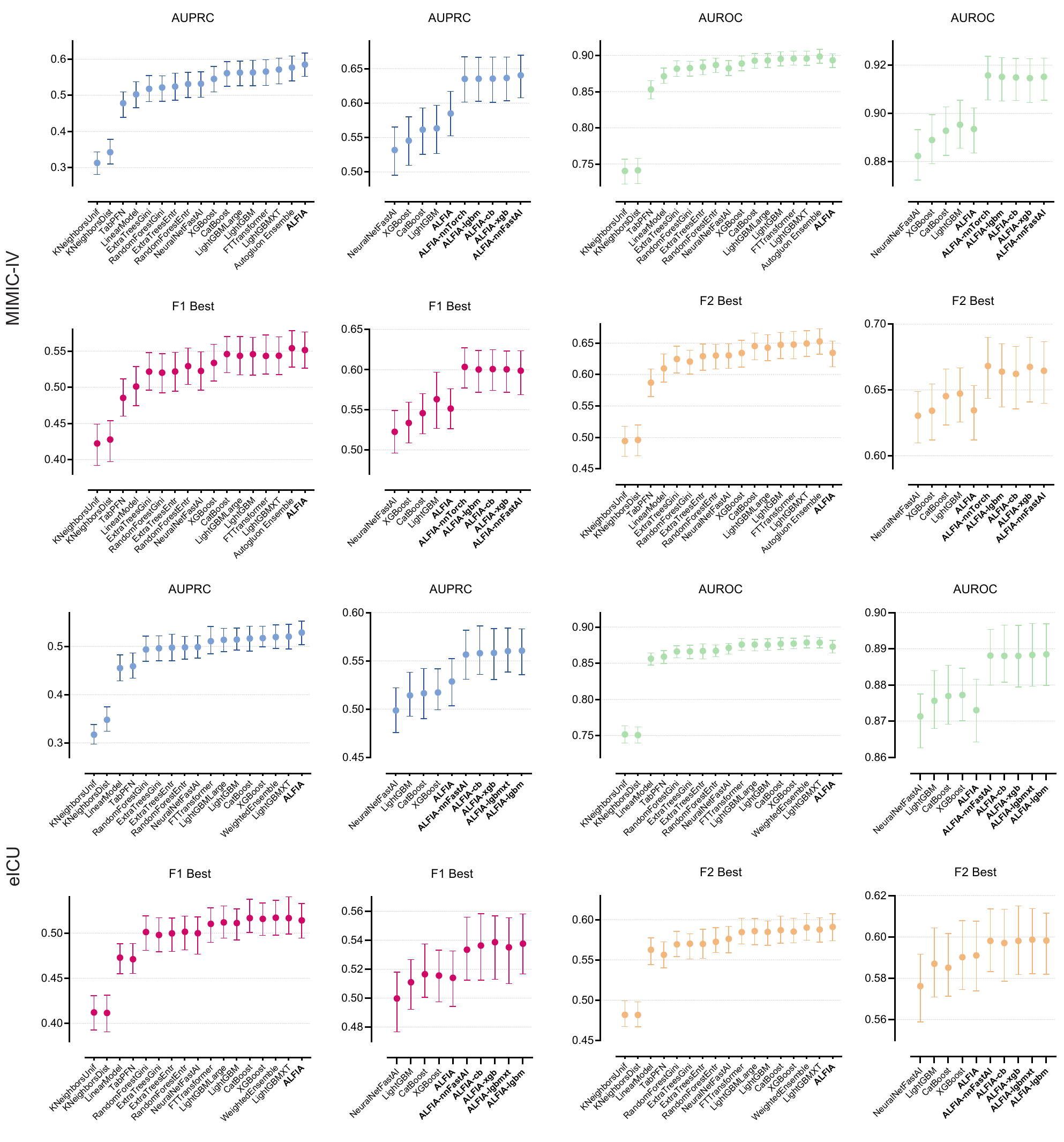} 
			\caption{\footnotesize \textbf{Figure 4. Performance comparison of machine learning models on MIMIC-IV and eICU datasets across multiple evaluation metrics.} The figure presents box plots comparing the performance of various machine learning algorithms including traditional methods (KNeighbors, Linear Model, Random Forest, Extra Trees), advanced ensemble methods (XGBoost, CatBoost, LightGBM), neural networks (NeuralNetFastAI, FT-Transformer), and the proposed ALFIA method with its variants. Performance is evaluated using four metrics: AUPRC (Area Under the Precision-Recall Curve), AUROC (Area Under the Receiver Operating Characteristic Curve), F1 Best, and F2 Best scores. The up panels show results for the MIMIC-IV dataset, while the right panels display results for the eICU dataset.}
			\label{fig:figure4} 
		\end{figure}

		According to our experimental results in test sets (\hyperref[tab:table3]{Table 3} and \hyperref[fig:figure4]{Figure 4}), ALFIA consistently outperforms baseline approaches in AUPRC across both the MIMIC-IV and eICU datasets, with improvements ranging from 0.5 to 1 percentage points. Interestingly, our model outperforms the Autogluon ensemble technique. ALFIA also maintains competitive F1 scores on the MIMIC dataset while demonstrating outstanding F2 scores on the eICU dataset.
		
		To confirm our technique, we ran additional experiments that combined the embeddings generated by the ALF module from training, validation, and test sets with original features and fed them into mainstream machine learning algorithms for further training. This effectively replaces the previous attention-based classification head. This practice resulted in significant performance increases, with gains of around 2-3 percentage points over baseline procedures.
		These thorough experimental results lead us to the conclusion that our suggested model successfully learns meaningful representations from data and produces considerable performance improvements across both benchmark datasets, confirming our approach's effectiveness and generalizability.
		
		\end{multicols}
		
		\captionsetup{
			justification=raggedright,  
			singlelinecheck=false,      
			skip=3pt                    
		}
		\footnotesize 
		
		\begin{longtable}{@{}
				>{\RaggedRight}p{0.32\linewidth} 
				*{4}{>{\centering\arraybackslash}p{\dimexpr (\linewidth - 0.32\linewidth - 8\tabcolsep)/4\relax}} 
				@{}}
			\caption{\textbf{Table3.} Performance Comparison of Models on PRAUC, AUROC, Best F1, and Best F2 Metrics}
			\label{tab:table3} \\
			
			\toprule
			\textbf{Model} & \textbf{PRAUC} & \textbf{AUROC} & \textbf{\shortstack{F1 Score\\(best)}} & \textbf{\shortstack{F2 Score\\(best)}} \\
			\midrule
			\endfirsthead 
			
			\toprule
			\textbf{Model} & \textbf{PRAUC} & \textbf{AUROC} & \textbf{\shortstack{F1 Score\\(best)}} & \textbf{\shortstack{F2 Score\\(best)}} \\
			\midrule
			\multicolumn{5}{@{}l}{\footnotesize\textit{Table \ref{tab:table3} continued from previous page}} \\ 
			\endhead
			
			\midrule
			\multicolumn{5}{r}{\footnotesize\textit{Continued on next page}} \\
			\endfoot
			
			\bottomrule
			\endlastfoot
			
			KNeighborsUnif   & 0.313 (0.282-0.343) & 0.741 (0.722-0.757) & 0.423 (0.392-0.449) & 0.495 (0.470-0.518) \\
			KNeighborsDist   & 0.343 (0.310-0.378) & 0.741 (0.723-0.758) & 0.428 (0.397-0.454) & 0.496 (0.471-0.520) \\
			TabPFN           & 0.478 (0.439-0.510) & 0.853 (0.840-0.865) & 0.485 (0.460-0.512) & 0.587 (0.565-0.609) \\
			LinearModel      & 0.503 (0.466-0.538) & 0.871 (0.862-0.883) & 0.501 (0.475-0.529) & 0.610 (0.588-0.633) \\
			ExtraTreesGini   & 0.518 (0.483-0.555) & 0.882 (0.871-0.893) & 0.522 (0.496-0.548) & 0.625 (0.603-0.645) \\
			RandomForestGini & 0.522 (0.483-0.554) & 0.883 (0.872-0.892) & 0.520 (0.493-0.547) & 0.621 (0.601-0.639) \\
			ExtraTreesEntr   & 0.524 (0.487-0.562) & 0.884 (0.873-0.894) & 0.522 (0.494-0.548) & 0.629 (0.607-0.649) \\
			RandomForestEntr & 0.531 (0.494-0.564) & 0.887 (0.877-0.896) & 0.529 (0.504-0.554) & 0.630 (0.609-0.648) \\
			NeuralNetFastAI  & 0.532 (0.495-0.565) & 0.882 (0.872-0.893) & 0.523 (0.496-0.549) & 0.631 (0.610-0.649) \\
			XGBoost          & 0.545 (0.510-0.580) & 0.889 (0.879-0.899) & 0.534 (0.509-0.559) & 0.634 (0.612-0.655) \\
			CatBoost         & 0.561 (0.525-0.593) & 0.893 (0.883-0.903) & 0.546 (0.520-0.570) & 0.645 (0.623-0.666) \\
			LightGBMLarge    & 0.563 (0.527-0.595) & 0.893 (0.884-0.903) & 0.544 (0.517-0.570) & 0.643 (0.622-0.663) \\
			LightGBM         & 0.563 (0.527-0.597) & 0.895 (0.886-0.906) & 0.546 (0.517-0.569) & 0.647 (0.626-0.667) \\
			\textbf{FTTransformer}    & 0.566 (0.527-0.599) & \textbf{0.896 (0.887-0.906)} & 0.543 (0.518-0.572) & 0.648 (0.625-0.668) \\
			\textbf{LightGBMXT}       & 0.571 (0.532-0.603) & 0.896 (0.886-0.906) & 0.544 (0.518-0.570) & \textbf{0.649 (0.629-0.670)} \\
			\textbf{Autogluon Ensemble} & 0.577 (0.540-0.609) & \textit{{*0.899 (0.890-0.909)}} & \textit{{*0.554 (0.528-0.578)}} & \textit{{*0.653 (0.632-0.673)}} \\
			\textbf{ALFIA} & \textbf{0.585 (0.552-0.617)} & 0.894 (0.884-0.902) & \textbf{0.552 (0.526-0.576)} & 0.635 (0.612-0.653) \\
		\end{longtable}
		
		\begin{multicols}{2}
		\subsection{ALFIA Exhibits Impressive Reasoning Capabilities}
\begin{figure}[htbp] 
			\centering
			\includegraphics[width=1\linewidth]{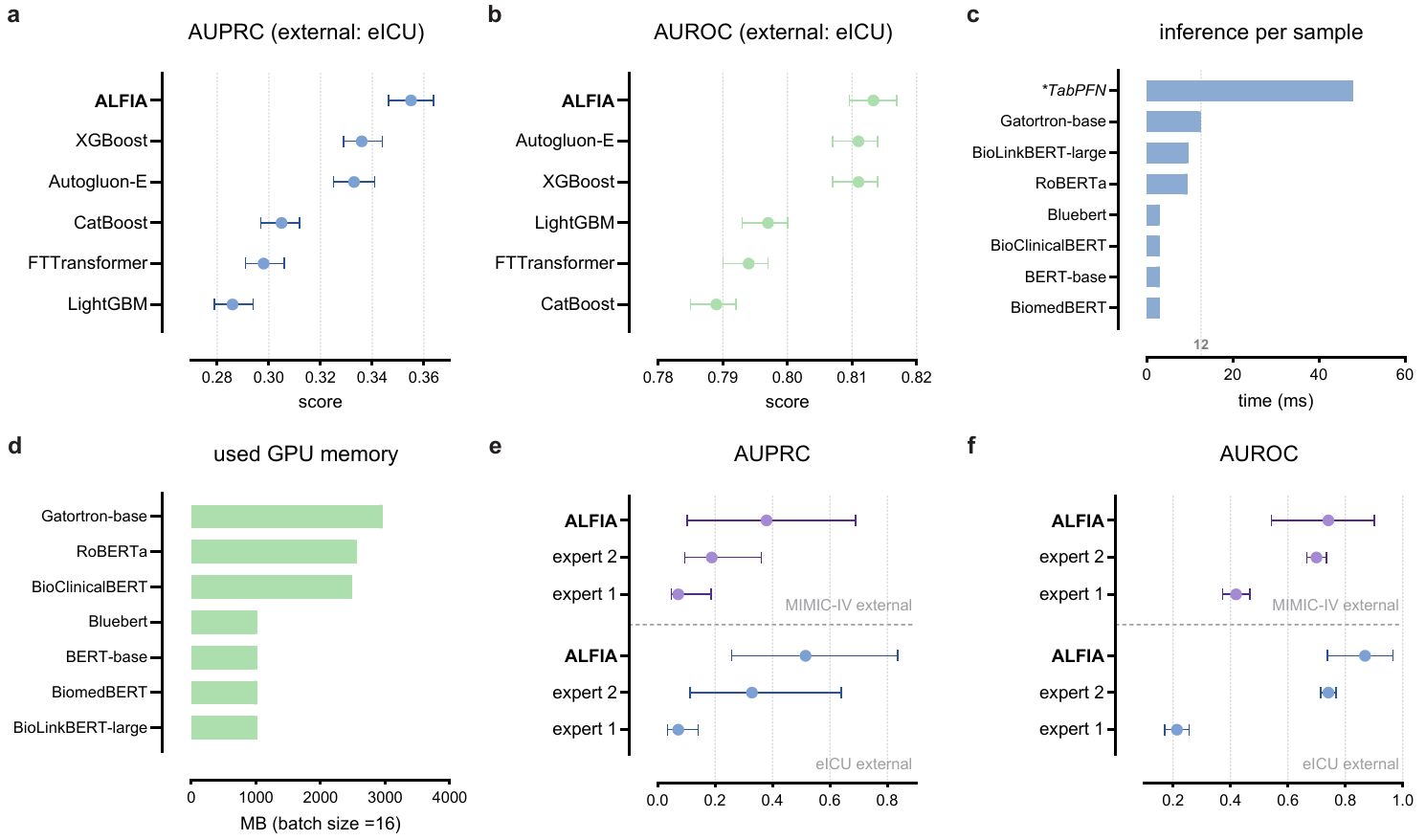} 
			\caption{\footnotesize \textbf{Figure 5. Performance evaluation of ALFIA on external eICU dataset.} \textbf{(a)} AUPRC scores comparing ALFIA against baseline methods including XGBoost, AutoGluon Ensemble, CatBoost, FT-Transformer, and LightGBM on the external eICU validation set. \textbf{(b)} AUROC scores for the same model comparison. \textbf{(c)} Inference time per sample (ms) for different BERT-based models and TabPFN on RTX4090 GPU.  \textbf{(d)} GPU memory consumption (MB) for different BERT base models under batch size 16 and maximum token length 512, ranging from 1-3 GB depending on model size. \textbf{(e)} Comparison between ALFIA and ICU expert scores in AUPRC with 95\% CI. \textbf{(f)} Comparison between ALFIA and ICU expert scores in AUROC with 95\% CI.}
			\label{fig:figure5} 
		\end{figure}

		\setlength{\parskip}{0.3em}
		\normalsize
		To confirm ALFIA's better reasoning generalization capabilities, we performed a detailed inference evaluation on the aforementioned cw-24 benchmark standard eICU dataset (n=150,000). We maintained and mapped the intersecting features between the eICU and MIMIC datasets, and we performed mode imputation on characteristics that were present in MIMIC but not in eICU. The results show that ALFIA outperforms GBDTs and the attention-based FT-Transformer, with gains of about 1.5 percentage points in AUPRC and 0.5 percentage points in AUROC (\hyperref[fig:figure5]{Figure 5a, b}).
		
		We looked more into ALFIA's hardware performance requirements. We ran inference experiments using several base BERT models on the eICU mapping dataset to investigate inference speed, comparing it to TabPFN, which also uses Transformer decoders. Our findings show that on RTX4090, regardless of the base model, the average inference time stays less than TabPFN's 48ms per sample, with the quickest achieving roughly 3ms per sample and the best-performing BioLinkBERT requiring 8-9ms per sample (\hyperref[fig:figure5]{Figure 5c}). This may be due to TabPFN's limits in large-sample inference, but it also demonstrates our model's outstanding inference speed performance. In terms of memory needs, the HuggingFace package default settings with batch size 16 and maximum token length 512 result in memory consumption ranging from 1-3GB depending on base model size (\hyperref[fig:figure5]{Figure 5d}). These specifications can be met by contemporary mainstream mid-to-low-end graphics cards.

		Furthermore, comparisons with professionals about inference revealed the superiority of ALFIA. We enlisted two specialists from the Intensive Care Unit of Guangdong Medical University Affiliated Hospital to partake in the comparison. For this assessment, we selected 100 novel instances from each of the MIMIC-IV and eICU datasets, allowing the model trained on MIMIC-IV to infer eICU cases and vice versa, and requested evaluations from both the model and experts. The results indicated that our model surpassed clinical specialists in previously unencountered cases in both AUPRC and AUROC (\hyperref[fig:figure5]{Figure 5e, f}), further evidencing its superior generalization and inference ability.
		\end{multicols}
		
		\clearpage 
		
		\begin{multicols}{2}
		
		\subsection{ALFIA improves classification performance by optimizing the latent space distribution of samples}
\begin{figure}[htbp] 
		\centering
		\includegraphics[width=1\linewidth]{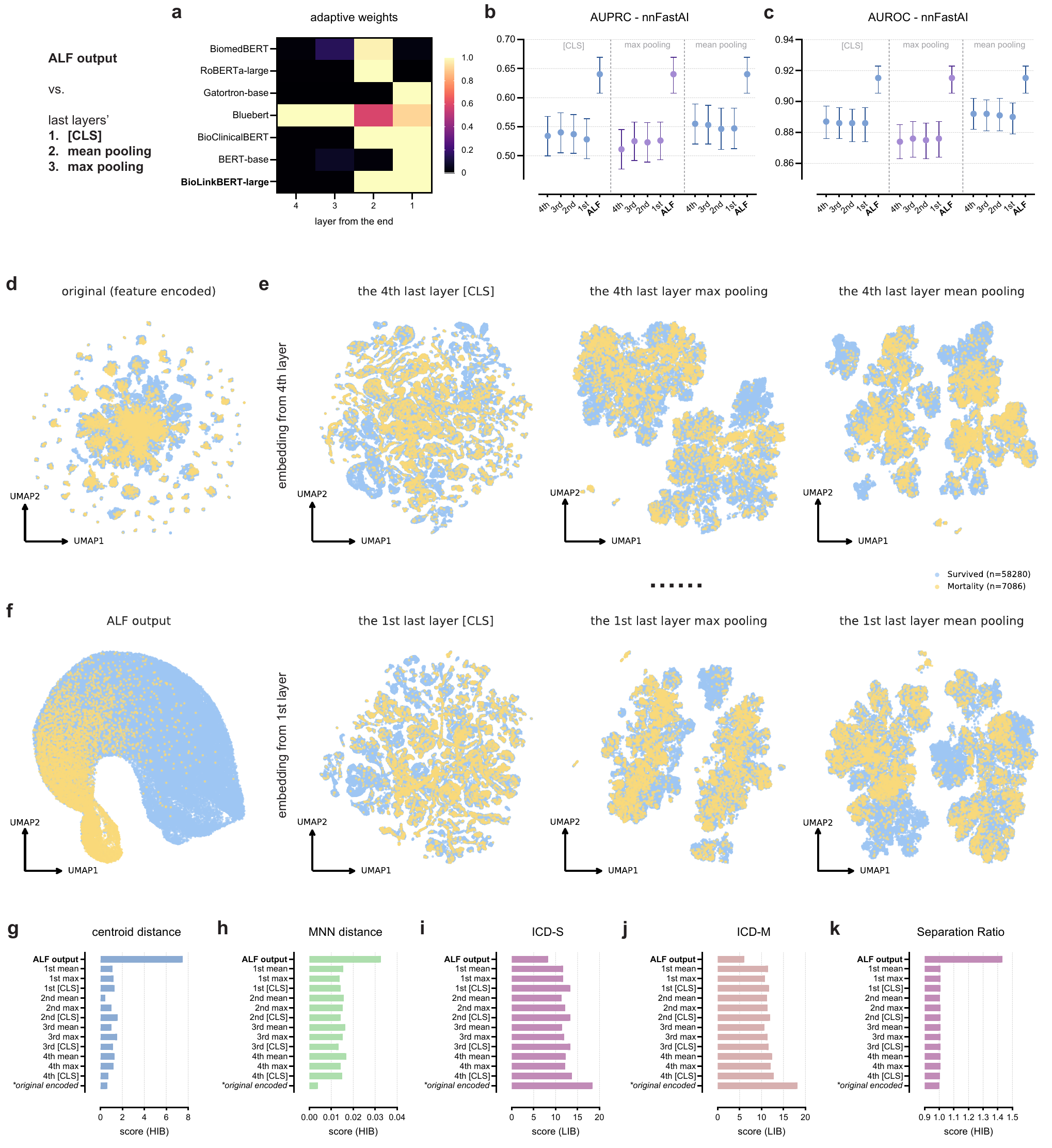} 
		\caption{\footnotesize \textbf{Figure 6. ALFIA optimizes latent space distribution of samples.} \textbf{(a)} Heatmap of layer attention weights across different BERT models. \textbf{(b)} Comparison of nnFastAI AUPRC performance (with 95\% CI) between BERT last four layers embedding methods (CLS, max, mean pooling) and ALF output. \textbf{(c)} Comparison of nnFastAI AUROC performance (with 95\% CI) between BERT last four layers embedding methods (CLS, max, mean pooling) and ALF output. \textbf{(d)} UMAP dimensionality reduction plot of original tabular matrix after feature encoding. \textbf{(e)} UMAP dimensionality reduction plot of sample matrices using BERT last four layers embedding methods (CLS, max, mean pooling). \textbf{(f)} UMAP dimensionality reduction plot of sample matrix from ALF output embeddings. \textbf{(g)} Bar chart of centroid distances across different methods. \textbf{(h)} Bar chart of average minimum neighbor distances across different methods. \textbf{(i)} Bar chart of intra-group distances for survival group across different methods. \textbf{(j)} Bar chart of intra-group distances for mortality group across different methods. \textbf{(k)} Bar chart of separation distances across different methods. In all UMAP plots, blue represents survival samples and yellow represents mortality samples. HIB: higher is better; LIB: lower is better.}
		\label{fig:figure6} 
		\end{figure}

		\setlength{\parskip}{0.3em}
		\normalsize
		To study what ALFIA does to the embedding of clinical assertions, we did a preliminary in-depth exploration using latent vector space. The ALF module, as conceived, accomplishes feature fusion using training layer weights, which are learned across distinct backbone models (\hyperref[fig:figure6]{Figure 6a}). Furthermore, we collected CLS, max pooling, and mean pooling from different encoder layers of the BioLinkBERT model and used them as feature inputs for the nnFastAI model in AutoGluon, comparing them to the embeddings given by ALF, which outperformed the nnFastAI model (see \hyperref[fig:figure4]{Figures 4}). CLS, max pooling, and mean pooling all demonstrated considerable performance stratification when compared to the ALF output embeddings (\hyperref[fig:figure6]{Figures 6b, c}), as measured by both AUPRC and AUROC.
		
		To better understand the latent space distribution of samples under different processing methods, we reduced the sample matrices' dimensionality to two dimensions using UMAP (the original feature engineering matrix processed through standard AutoGluon had 133 dimensions, while other BERT-encoded models had 1024 dimensions). It is obvious that neither the original distribution nor the derived inter-layer embeddings revealed unique distributional heterogeneity across in-hospital survival and mortality samples, but rather mutual fusion and overlap (\hyperref[fig:figure6]{Figures 6d, e}). In contrast, our ALF output clearly exhibited discrete high-density regions for mortality and survival samples, as well as their transition zones (\hyperref[fig:figure6]{Figure 6f}), implying that ALF output represents a more task-optimized latent space distribution. We measured the latent space features of samples and discovered that ALF output embeddings performed best across all measures, including inter-group centroid distance, average minimum neighbor distance, intra-group distance, and classification degree ratio (\hyperref[fig:figure6]{Figures 6g-k}).
		
		Finally, all results show that our proposed architecture ALFIA and its extensions ALFIA-boost/nn outperform other architectures in classification while keeping generalization capability.
		
		\end{multicols}
		
		\begin{multicols}{2}
		\normalsize
		\section{Discussion}
		\vspace{-0.8em}   
		\setlength{\parskip}{0.3em}
		ALFIA (Adaptive Layer Fusion with Intelligent Attention) is a cutting-edge deep learning architecture for text-based clinical prediction. We discovered that ALFIA outperforms state-of-the-art tabular classifiers and conventional machine learning algorithms on a variety of assessment criteria while maintaining robust generalization on external validation datasets.
		
		ALFIA's core innovation is adaptive layer fusion, which dynamically combines multiple-layer semantic representations from pre-trained Transformer models. On the MIMIC-IV dataset, ALFIA had a significantly higher AUPRC (0.585) than the AutoGluon ensemble (0.577) and the FT-Transformer (0.566). Given the class imbalance in mortality prediction tasks, AUPRC is a more reliable performance metric than AUROC, so this improvement is significant. Cross-dataset generalization demonstrates that models trained on data from a single institution may be deployed in numerous healthcare environments without losing performance.
		
		Specifically, our evaluation study compared ALFIA to clinical professionals. ALFIA surpassed ICU specialists in new case AUPRC and AUROC scores, indicating that it could be an effective clinical decision support tool. These findings demonstrate that the model can enhance clinical expertise, particularly in complex decision-making circumstances requiring elements outside the model's capabilities.
		
		The UMAP visualization demonstrated how ALFIA improves clinical data presenting. ALFIA's adaptive layer fusion generates high-density zones for several outcome categories, as opposed to other embedding methods that overlap survival and mortality cases. Our tests demonstrated improved classification performance due to latent space separability, as judged by inter-group centroid distance and silhouette scores. The trained layer attention weights of multiple BERT models reveal interesting patterns in how the model prioritizes representational layer input. In clinical applications, understanding the model's decision-making process increases clinician trust and assures proper model use.
		
		Several constraints should be addressed when interpreting our findings. First, our research relies on two large datasets (MIMIC-IV and eICU) from similar healthcare systems, which may not fully reflect global clinical practice. Future validation studies should include datasets from other fields and healthcare systems to improve generalizability.
		
		Second, our implementation only captures clinical data within the first 24 hours of ICU admission, potentially missing crucial temporal dynamics later in patient stays. ALFIA may involve longitudinal modeling to account for changes in patient state and treatment response over time.
		
		ALFIA's improved performance and computational efficiency suggest that it has a high clinical potential as an early warning system for ICU mortality. The model's ability to read routine clinical language without data preprocessing makes it perfect for EHR integration. Aside from technical performance, successful clinical implementation necessitates establishing alert thresholds to reduce false positives, designing user interfaces that communicate risk predictions to clinical staff, and implementing robust monitoring systems to detect model drift or performance degradation over time.
		
		Our research on this model design is preliminary; more work is required. Our data suggest several intriguing study choices. ALFIA's modular architecture, particularly its ability to combine ALF embeddings with gradient boosting methods (ALFIA-boost) and deep neural networks (ALFIA-nn), suggests ensemble approaches that employ a variety of complementary modeling techniques.
		
		Domain-specific pre-trained models (e.g., BioLinkBERT \cite{yasunaga_linkbert_2022}) achieved success in our experiments, highlighting the importance of developing language models for the medical domain. Future research can create generalized pre-trained models specifically for medical prediction tasks to improve the accuracy of healthcare predictions.
		
		The adaptive layer fusion approach proposed for ALFIA can be applied to various clinical prediction tasks, including length of stay estimation, readmission risk assessment, and adverse event prediction, and can even be extended to non-medical domains. Examining the transferability of our method across numerous similar scenarios will underscore the broad applicability of these architectural innovations.
			
		\normalsize
		\section{Conclusion}
		In conclusion, ALFIA represents a significant advancement in the application of deep learning to clinical mortality prediction. The model's superior performance, robust generalization capabilities, and computational efficiency position it as a promising tool for enhancing clinical decision-making in ICU settings. While important limitations and implementation challenges remain, our findings provide strong evidence for the potential of adaptive layer fusion approaches in medical AI applications and establish a foundation for future research in this critical area of healthcare technology.
		
		\section*{Author Contributions}
		H.W. conceptualized and designed the ALFIA model framework, developed the complete experimental pipeline, performed all model training and experimental evaluations, created data visualizations, and drafted the manuscript. 
		
		C.T. provided project supervision, experimental infrastructure, and computational resources.
		
		\section*{Data and Code Availability}
		The benchmark dataset used in this study is publicly available at \url{https://github.com/Hanziwww/CW-24}. The model implementation and article-related code can be accessed at \url{https://github.com/Hanziwww/ALFIA}. Publicly available data are deposited in Zenodo at \url{https://zenodo.org/records/15574378}.
		
		The MIMIC-IV dataset is available through PhysioNet at \url{https://physionet.org/content/mimiciv/3.1/} upon completion of required training and approval process. Access to the eICU Collaborative Research Database can be obtained at \url{https://physionet.org/content/eicu-crd/2.0/} following the same credentialing requirements as MIMIC-IV.
		
		Both clinical datasets require researchers to complete the CITI "Data or Specimens Only Research" training course and sign a data use agreement before gaining access. Detailed instructions for accessing these datasets are provided on the respective PhysioNet pages.
		
		\section*{Acknowledgements} 
		We acknowledge the contributions of various open-source projects utilized in this study and express our gratitude to the research teams behind the pre-trained BERT models cited in this work for their valuable contributions to the community. 
		
		We also extend our sincere appreciation to the MIT Laboratory for Computational Physiology for providing the MIMIC-IV database and to the Philips Healthcare team for making the eICU Collaborative Research Database publicly available. These critical care datasets have been instrumental in advancing research in healthcare informatics and clinical decision support systems.

		\printbibliography[title={References}]
		\end{multicols}
	\end{document}
	
	/*
	@book{knuth1984,
		author    = {Donald E. Knuth},
		title     = {The TeXbook},
		publisher = {Addison-Wesley},
		year      = {1984},
		series    = {Computers and Typesetting},
		volume    = {A}
	}
	
	@article{einstein1905,
		author    = {Albert Einstein},
		title     = {Zur Elektrodynamik bewegter K{\"o}rper},
		journal   = {Annalen der Physik},
		volume    = {322},
		number    = {10},
		pages     = {891--921},
		year      = {1905},
		DOI       = {10.1002/andp.19053221004}
	}
	
	@article{dirac1928,
		author    = {Paul Adrien Maurice Dirac},
		title     = {The Quantum Theory of the Electron},
		journal   = {Proceedings of the Royal Society of London. Series A, Containing Papers of a Mathematical and Physical Character},
		volume    = {117},
		number    = {778},
		pages     = {610--624},
		year      = {1928},
		DOI       = {10.1098/rspa.1928.0023}
	}
	
	@book{latexcompanion,
		author    = {Frank Mittelbach and Michel Goossens and Johannes Braams and David Carlisle and Chris Rowley},
		title     = {The LaTeX Companion},
		edition   = {2nd},
		publisher = {Addison-Wesley},
		year      = {2004}
	}
	*/